\documentclass[twoside,11pt]{article}

\usepackage{blindtext}
\usepackage[preprint]{style}
\usepackage{amsmath}

\usepackage{lastpage}

\firstpageno{1}

\begin{document}

\title{A free from local minima algorithm for training regressive MLP neural networks}

\author{\name Augusto MONTISCI \email augusto.montisci@unica.it \\
       \addr Electrical and Electronic Engineering Dept.\\
       University of Cagliari\\
       Italy}

\editor{My editor}

\maketitle

\begin{abstract}%
In this article an innovative method for training regressive MLP networks is presented, which is not subject to local minima. The Error-Back-Propagation algorithm, proposed by William-Hinton-Rummelhart, has had the merit of favouring the development of machine learning techniques, which has permeated every branch of research and technology since the mid-1980s. This extraordinary success is largely due to the black-box approach, but this same factor was also seen as a limitation, as soon more challenging problems were approached. One of the most critical aspects of the training algorithms was that of local minima of the loss function, typically the mean squared error of the output on the training set. In fact, as the most popular training algorithms are driven by the derivatives of the loss function, there is no possibility to evaluate if a reached minimum is local or global. The algorithm presented in this paper avoids the problem of local minima, as the training is based on the properties of the distribution of the training set, or better on its image internal to the neural network. The performance of the algorithm is shown for a well-known benchmark.
\end{abstract}


\section{Introduction}
Even if Machine Learning (ML) includes a multiciplity of paradigms, much different among them, most part can be considered as an evolution of Error Backpropagation Algorithm (EBP) of Multi Layer Perceptron (\cite{Rumelhart:86}). The merit of this algorithm consists in the fact that for the first time it was possible to train networks with an intermediate layer, and therefore reproduce non-linear input-output relationships. This was as true for classification problems as it was for regression problems. Concerning the latter ones, \cite{Cybenko:89} had demonstrated that a Perceptron with a single hidden layer is a universal approximator, however leaving open the problem of determining both the number of neurons needed to solve a specific problem, and how to determine the connection weights. The EBP algorithm offered a tool to determine the value of the parameters, while the determination of the optimal number of neurons is still an open problem. Successively,  methods have been presented to address both questions (\cite{Delogu:08}, \cite{Ploj:14}, \cite{Ploj:14b}, \cite{ Fernández-Delgado:14}, \cite{Fernández-Delgado:11}, \cite{Ploj:11}, \cite{Curteanu:11}, \cite{Carcangiu:09}), but the EBP paradigm, with an important set of variations, still represents the standard of machine learning. This paradigm consists in finding the minimum of a loss function, which is typically given by the output mean squared error with respect to the target value. All the minimization techniques also developed in contexts other than that of ML have been proposed to solve this problem, but the standard is represented by the use of first and second order descent methods (\cite{marquardt:63}), in whose category the EBP itself falls. First order algorithms, such as EBP, have made a comeback with the advent of Deep Learning (\cite{lecun:15}), as the huge number of parameters makes second order methods impractical, even in cases in which approximate expressions of the Hessian are used. The methods based on the derivatives of the loss function have the advantage of being simple to implement, but lack a criterion that allows to establish whether a stationarity point of the loss function represents a global minimum or not. However, there is another limitation that derives from this type of black-box approach. In fact, the fact of considering the loss function as the only indicator of the network performance leads to not distinguishing the different functions of the single parts of the network, as well as the relevance of single examples of the training set.\\
\begin{figure}[hbt!]
\centering
\includegraphics[width=0.75\linewidth]{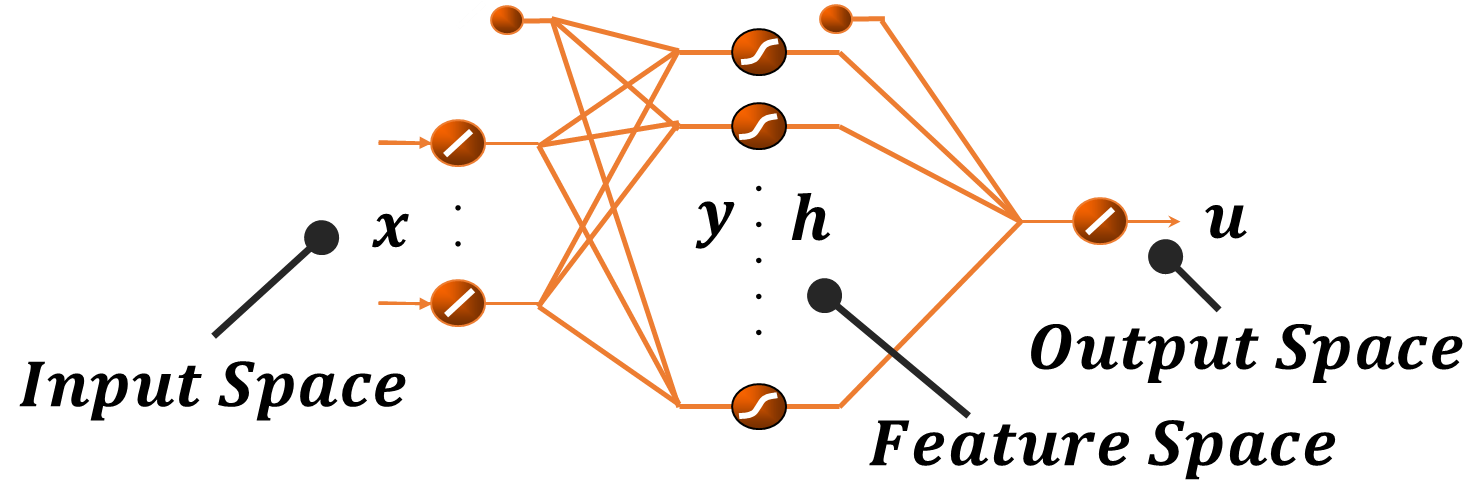}
\caption{Multilayer Perceptron scheme}
\label{fig: MLPMISO}
\end{figure}
As a premise to the description of the algorithm that will be presented in the next sections, it is worth rethinking the structure of the MLP network, going beyond the black-box interpretation. First of all, the need for the hidden layer derives from the fact that the input-output relationship is nonlinear. If this were not the case, a linear hidden layer would be completely useless, because the cascade of the two connecting layers would be completely equivalent to their product. Furthermore, since the algebraic relationship between the output of the hidden layer, called \textit{Feature Space} (\cite{Cortes:95}) and the output of the neural network is linear, we can think that the function of the hidden layer is to make linear the relationship with the output. In other words, for each output neuron, the points of the training set in the product space $\{$feature space$\} \times \{$output space$\}$ must be coplanar. In fact, any residual nonlinearity downstream of the feature space could not be corrected by the last layer of connections. Hence also the fact that there is no reason to include in the loss function the weights of the connections preceding the output, since given the points in the feature space, the optimal set of weights is the one that corresponds to the linear regression plane. Another important consideration concerns the different role of the hidden layer and the output layer. Indeed, if the former represents the degrees of freedom of the network, because as the size of the feature space increases, it becomes easier to make the points of the training set coplanar in the product space $\{$feature space$\} \times \{$output space$\}$, the latter constitutes a constraint, because the same distribution in the feature space must satisfy the coplanarity constraint with respect to different outputs.

This paper presents a training algorithm based on this interpretation of the algebraic structure of the MLP. For sake of simplicity and without prejudice to the general validity of the results, in this article the case of regressive MLP networks with only one output (MISO) will be treated. The paper is structured as follows: in the first section, the analytical basis of the procedure is presented. The second section briefly describes the choices adopted for implementation in the Matlab environment. The third shows the results obtained with a well-known benchmark. At the end, conclusions are provided.

\section{Description of the algorithm}

\begin{equation}
\begin{cases}
 \mathbf{ W \cdot x + d = y} \\
\mathbf{ h = \sigma(y)} \\
\mathbf{ V \cdot h + b = u}
\end{cases} 
\label{equ: MLP}
\end{equation}

In this section the training algorithm will be described, referring to the scheme shown in Fig.~\ref{fig: MLPMISO}. The choice of having only one output serves to simplify the discussion, but does not affect the generality of the problem. The MLP network implements an algebraic structure represented by the system of equations ({\ref{equ: MLP}), where $\boldsymbol{x}$ is the input to the network, $\boldsymbol{W}$ is the weight matrix of the first connection layer, $\boldsymbol{d}$ is the bias of the first layer, $\boldsymbol{y}$ is the input vector to the hidden layer, $\sigma$ is the activation function of the hidden layer, $ \boldsymbol{h}$ is the image of the input in the feature space, $\boldsymbol{V}$ is the vector of the weights of the connections with the output, $\boldsymbol{b}$ is the bias of the second layer. The first layer of connections linearly transforms the points from the input space of the network into the input space of the hidden layer. The hidden layer activation function is the only nonlinearity of the network. Through the first two transformations of the ({\ref{equ: MLP}), the distribution of the training set points in the input space is transformed into a distribution in the feature space. The first layer of connections must ensure that the points of the training set in the product space $\{$feature space$\} \times \{$output space$\}$ are all coplanar. If this does not happen, the second layer of connections will not be able to compensate for these errors. The solution that minimizes least squares is the regression plan. For this reason, even if a conventional training algorithm is applied, it is not convenient to include the weights of the second layer of connections within the loss function. In fact, it is better to define the loss as a function of the weights of the first layer and, by freezing this, to define the second layer by calculating the regression plane.

The algorithm presented in this paper, instead of finalizing the search to minimize a loss function, aims to construct the feature space in such a way as to ensure the coplanarity of the points, or equivalently that the level curves of the function in the input space are transformed into straight lines in the feature space (Fig.~\ref{fig: Linearization}).

\begin{figure}[hbt!]
\centering
\includegraphics[width=0.75\linewidth]{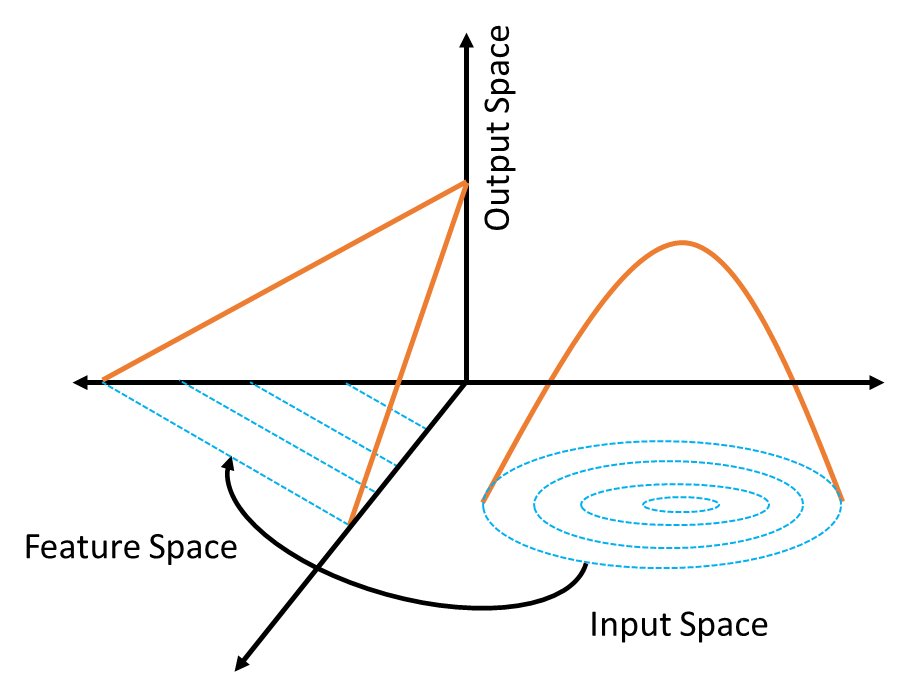}
\caption{Linearizzation performed by the hidden layer}
\label{fig: Linearization}
\end{figure}

The condition for linearization can be expressed analytically as follows:

\begin{equation}
\mathbf{ s^T \cdot \sigma(W \cdot x_i + d)} = \textit{a}\cdot \mathbf{u_i} \quad \forall \textit{i}=1 \dots \textit{N}
\label{equ: Lines Family}
\end{equation}
where $\boldsymbol{s}$ is a vector of linear combination coefficients of the points in the feature space, while $a$ is a scalar of arbitrary value, which determines the distance between the straight lines of level, and consequently the slope of the regression plan. Finally, $N$ is the number of points in the training set. Dividing the equation ({\ref{equ: Lines Family}) by $a$ eliminates the uncertainty of the solution:

\begin{equation}
\mathbf{ \hat{s}^T \cdot \sigma(W \cdot x_i + d)} = \mathbf{u_i} \quad \forall \textit{i}=1 \dots \textit{N}
\label{equ: Lines Family Norm}
\end{equation}
where $\hat{\boldsymbol{s}}$ is the normalized variable. Globally the equation ({\ref{equ: Lines Family Norm}) is a system of $N$ nonlinear equations in the variables $\boldsymbol{W}$, $\boldsymbol{d}$, $\boldsymbol{\hat{ s}}$. Fixing the values of $\boldsymbol{W}$ and $\boldsymbol{d}$ the same system becomes a system of linear equations overdetermined in the single variable $\hat{\boldsymbol{s}}$.

The proposed algorithm starts from a first order approximation of the first member of the equation ({\ref{equ: Lines Family Norm}). Let $\boldsymbol{W}_0$, $\boldsymbol{d}_0$, $\boldsymbol{\hat{s}}_0$ be an initial solution of the system ({\ref{equ: Lines Family Norm}), and $\delta\boldsymbol{W}$, $\delta\boldsymbol{d}$, $\delta\boldsymbol{\hat{s}}$ the respective increments of the three variables that provide the solution:

\begin{equation}
\mathbf{ (\hat{s}_0+\delta s)^T \cdot \sigma[(W_0+\delta W) \cdot x_i + (d_0+\delta d)]} = \mathbf{u_i} \quad \forall \textit{i}=1 \dots \textit{N}
\label{equ: SystemIncrem}
\end{equation}

Now replace the left-hand side of the equation with a first-order approximation of the incremented function:

\begin{equation}
\mathbf{ (\hat{s}_0+\delta s)^T \cdot [\sigma_0 + \nabla \sigma(W,d) \odot \delta_{W,d} \cdot 1]} \approx \mathbf{u_i} \quad \forall \textit{i}=1 \dots \textit{N}
\label{equ: System 1stOrd}
\end{equation}
where $\boldsymbol{\sigma_0}$ is the vector in the feature space corresponding to a zero increment, $\nabla \sigma(\boldsymbol{W,d})$ is the gradient calculated with respect to the weights of the first layer connections, $ \delta_{\boldsymbol{W,d}}$ is the increment of the variables, $\odot$ is the elementwise multiplication, $\boldsymbol{1}$ is a column vector of only 1s. Reordering the equation and neglecting the infinitesimals of higher order we get:

\begin{equation}
\mathbf{ \sigma_0^T \cdot \delta s + \hat{s}_0^T \cdot  \nabla \sigma(W,d) \odot \delta_{W,d} \cdot 1}  \approx \mathbf{u_i-\hat{s}_0^T \cdot \sigma_0} \quad \forall \textit{i}=1 \dots \textit{N}
\label{equ: SystemReordered}
\end{equation}

The equation ({\ref{equ: SystemReordered}) defines a linear, overdetermined system of equations, whose unknowns are the increments of the parameters. By combining ({\ref{equ: Lines Family Norm}) with ({\ref{equ: SystemReordered}) it is possible to implement an iterative procedure which provides the weights of the first layer of connections. Taking into account the fact, as previously indicated, that the second layer of connections is uniquely determined once the weights of the first layer have been assigned, the iterative procedure actually completes the training procedure.

The procedure takes place as follows. Given the initial value of the weights of the first layer of connections, the system ({\ref{equ: Lines Family Norm}) is resolved into the single variable $\boldsymbol{\hat{s}}$. Since the system is overdetermined, it can only be solved by minimizing the root mean square error (\cite{jodar:91}). Using the obtained solution, together with the same weights of the first layer used previously, the coefficients and the known term of the system ({\ref{equ: SystemReordered}) are obtained. This system is also overdetermined, and therefore it too will have to be solved according to least squares. Only the increments of the weights of the first layer $\delta_{\boldsymbol{W,d}}$ of the solution of the system are used, while $\boldsymbol{\hat{s}}$ is updated by solving the system ({\ref{equ: Lines Family Norm}) after updating the coefficients based on the new values of $\mathbf{W}$ and $\mathbf{d}$. The iterative process ends when the error in the system ({\ref{equ: Lines Family Norm}) is less than a pre-set threshold value. Alternatively, since the second layer of weights are uniquely determined by the first layer of weights, the error value of the current solution can be computed directly on the neural network output.
Since a first-order approximation is being used, the question remains open as to what is the appropriate step length to take at each iteration. For this reason, the Line Search method is adopted (\cite{bazaraa:13}), sampling the segment from the current solution to the increased one, and evaluating for each point the value of the output error. At each iteration, the assigned increment will be the one that corresponds to the minimum error.

As can be deduced, the number of operations to be performed for each iteration is much higher than that of a conventional method, but the proposed method has the advantage of not being subject to local minima problems, since the objective is not more that of minimizing a loss function, rather than solving a system of equations, and therefore the loss function that drives the search for the solution coincides with the goal of the training, rather than, as usually happens, being constituted by an aggregate target function.

It is worth noting that the proposed method, precisely because it is not based on the minimization of an aggregate loss function, allows the use of different criteria in the application of the line search, such as for example the minimization of the maximum error in the training set. The complete analysis of the potential of the method goes beyond the objectives of this paper and will be the subject of future developments.

\section{Results}

A well-known benchmark for testing optimization methods (\cite{schwefel:81}) is used as an application example. In cases where the search for the optimal solution is carried out with the aid of a neural network (\cite{Carcangiu:09b}), the accuracy of the network and its generalization capability assume fundamental importance. The general expression of the function is:

\begin{equation}
f(\textbf{x}) = 418.9829\cdot d - \sum_{i=1}^{d} x_i \sin(\sqrt{\lvert x_i \rvert})
\label{equ: Schwefel}
\end{equation}
where $d$ is the size of the space where the function is defined. In this case a dimension $d=3$ was used, and a range of variables [-500,500]. Fig.~\ref{fig: Schwefel} shows the Schwefel function used as a test. The value of the function is represented by the color gradient. It is apparent that the function has a large number of maxima and minima, which implies that a large number of hidden neurons will be required to adequately fit it.

\begin{figure}[hbt!]
\centering
\includegraphics[width=0.75\linewidth]{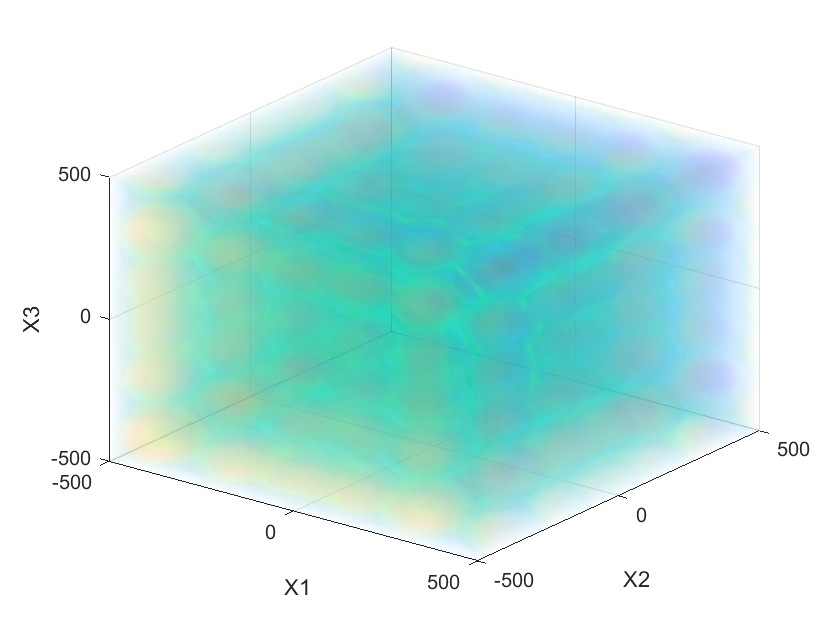}
\caption{Schwefel's function defined in \textit{d}=3}
\label{fig: Schwefel}
\end{figure}
 
An MLP network with 200 hidden nodes has been trained for 4000 epochs. Fig.~\ref{fig: Perf} shows the trend of the mean squared error during training. Note that the error in the diagram refers to the output data previously normalized between -1 and 1. The total number of training examples was equal to 5152. As can be seen, in the first iterations there is a considerable reduction of the error, while subsequently the descent speed is considerably reduced both in absolute and relative terms. In some segments it can be noted that the average error appears constant, and in some cases it can be slightly increasing, but this does not prevent, after a suitable number of iterations, from resuming the decreasing trend. This means that although the mean (squared) error is constant, in reality the iterative process is modifying the configuration of the weights, which allows the network to identify regions in the parameter space where it is possible to obtain performance improvements. As can be seen, the error value after 4000 epochs had not yet stabilized, although the decrease had become much slower.

As mentioned earlier, it is possible to redefine the search criterion based on the maximum error in the training set rather than the mean value, but this, like other issues affecting convergence acceleration, is beyond the scope of this paper and will be the subject of future works.

\begin{figure}[hbt!]
\centering
\includegraphics[width=0.75\linewidth]{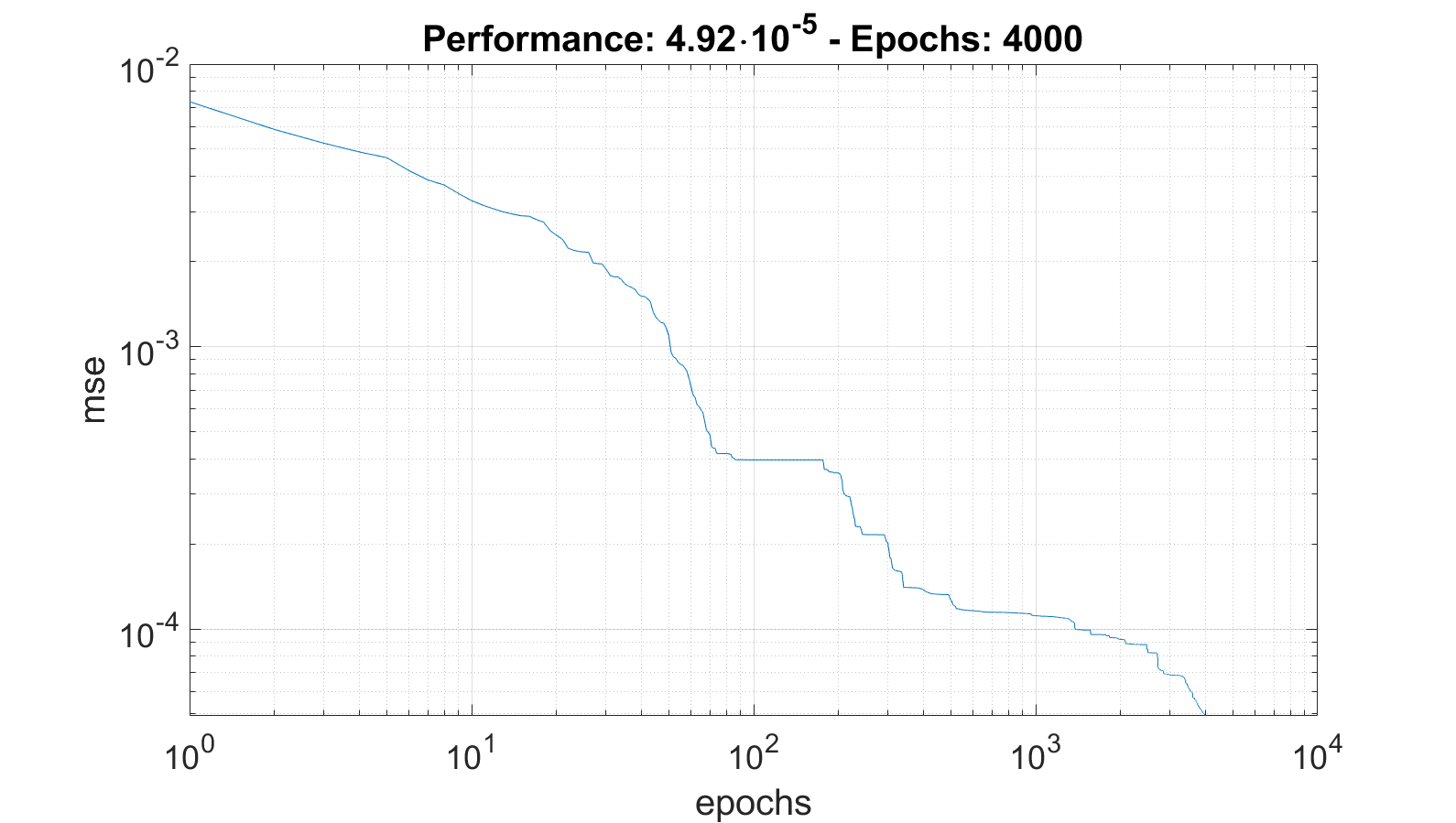}
\caption{Performance diagram of the training}
\label{fig: Perf}
\end{figure}

\section{Some implementation notes}
The algorithm was developed in Matlab environment. This has led, as far as possible, to exploit matrix calculation, rather than resorting to \textit{for} loops. This applies in particular to the construction, at each iteration, of the matrix of the system coefficients, which has made it possible to enormously contain the time required for a single iteration.

It is also worth making some considerations on the application of the Line Search, with which the length of the step to be taken at each iteration is established. The number of samples that are explored for each iteration is a macro-parameter that must be set by the designer, and significantly affects the calculation time of the single iteration. In the case under examination, the increment obtained as a solution of the linear system is divided into 1000 points distributed with a logarithmic law, so as to better detail the trend of the function for small increments. It may happen that the minimum of the error corresponds to the first sample, which implies that the minimum could correspond to a smaller increment than the minimum foreseen. In this case the search for the minimum is further refined using the bisection method applied to the interval between 0 and the smallest increment. If the minimum falls in correspondence with a zero increase, the training procedure stops.

The procedure requires solving a large number of linear systems. The use of the Matlab library functions in the case where the matrix of the coefficients is ill-conditioned involves difficulties, both in terms of calculation times and instability of the iterative process. For this reason, also taking into account that the system is approximated in its definition, we use a solution method based on successive projections (\cite{Cannas:12}), rather than on the inversion of the coefficient matrix. The obtained solution is anyway affected by the misconditioning of the coefficient matrix, and this generally results in a less precise solution, but the algorithm has the merit of quickly providing a reasonable solution.

Again regarding the problem of the misconditioning of coefficient matrices, it has been observed that a large number of examples of the training set helps to mitigate the problem. Increasing the training set does not significantly affect the calculation time, if not indirectly due to memory occupation. It has also been observed that the calculation time of the single iteration does not significantly depend on the number of hidden neurons, therefore it is observed that as the number of hidden neurons increases, the total training time decreases, because the number of epochs necessary to reach the target error is smaller.

As stopping criterion a double check was adopted on the number of epochs and on the minimum variation of the error between two successive epochs. In all the tested cases, irrespective of the number of examples, the algorithm stopped when the maximum number of epochs was reached. In future developments it is planned to use a stopping criterion based on the stabilization of the weights of the connections, rather than on the outputs. This will prevent the procedure from stopping due to a momentary stabilization of the error trend, like the one found in the example in the previous section. If, in fact, it is observed that the weights of the connections continue to change, even if this does not translate into a reduction in the average error of the network, the training must still proceed. Analyzing weight trends during training offers the possibility of accelerating learning. If, in fact, this trend is regular, it is generally possible to extrapolate the values to predict which value it is tending to. This possibility will be the subject of future developments.

\section{Conclusion}
This paper describes a training algorithm for Multi Layer Perceptron neural networks whose main feature is that it is not subject to the problem of local minima, typical of conventional algorithms, in which the search for the minimum of the loss function is driven by derivatives of the first and second order of the loss function. Even at a preliminary implementation, the algorithm shows good performance and limited computational complexity as both the training set and the network size increase, which makes it a possible candidate for the treatment of big data problems using neural networks. The performance of the algorithm has been tested on a benchmark that is difficult to treat with the most common training algorithms, obtaining very encouraging results. Finally, the paper introduces lines of development that will be the subject of future publications.

\vskip 0.2in
\bibliography{sample}

\begin{thebibliography}{18}
\providecommand{\natexlab}[1]{#1}
\providecommand{\url}[1]{\texttt{#1}}
\expandafter\ifx\csname urlstyle\endcsname\relax
  \providecommand{\doi}[1]{doi: #1}\else
  \providecommand{\doi}{doi: \begingroup \urlstyle{rm}\Url}\fi

\bibitem[Bazaraa et~al.(2013)Bazaraa, Sherali, and Shetty]{bazaraa:13}
M.S. Bazaraa, H.D. Sherali, and C.M. Shetty.
\newblock \emph{Nonlinear programming: theory and algorithms}.
\newblock John wiley \& sons, 2013.

\bibitem[Cannas et~al.(2012)Cannas, Carcangiu, Fanni, Forcinetti, Montisci,
  Concu, et~al.]{Cannas:12}
B.~Cannas, S.~Carcangiu, A.~Fanni, R.~Forcinetti, A.~Montisci, G.~Concu, et~al.
\newblock Hops: A new tomographic reconstruction algorithm for non destructive
  acoustic testing of concrete structures.
\newblock \emph{Advances in civil engineering and building materials},
  831:\penalty0 259--263, 2012.

\bibitem[Carcangiu et~al.(2009{\natexlab{a}})Carcangiu, Fanni, and
  Montisci]{Carcangiu:09}
S.~Carcangiu, A.~Fanni, and A.~Montisci.
\newblock A constructive algorithm of neural approximation models for
  optimization problems.
\newblock \emph{COMPEL-The international journal for computation and
  mathematics in electrical and electronic engineering}, 28\penalty0
  (5):\penalty0 1276--1289, 2009{\natexlab{a}}.

\bibitem[Carcangiu et~al.(2009{\natexlab{b}})Carcangiu, Fanni, and
  Montisci]{Carcangiu:09b}
S.~Carcangiu, A.~Fanni, and A.~Montisci.
\newblock \emph{Multi objective optimization algorithm based on neural networks
  inversion}, pages 744--751.
\newblock 2009{\natexlab{b}}.

\bibitem[Cortes and Vapnik(1995)]{Cortes:95}
C.~Cortes and V.~Vapnik.
\newblock Support-vector networks.
\newblock \emph{Machine Learning}, 20\penalty0 (3):\penalty0 273 – 297, 1995.

\bibitem[Curteanu and Cartwright(2011)]{Curteanu:11}
S.~Curteanu and H.~Cartwright.
\newblock Neural networks applied in chemistry. i. determination of the optimal
  topology of multilayer perceptron neural networks.
\newblock \emph{Journal of Chemometrics}, 25\penalty0 (10):\penalty0 527 –
  549, 2011.

\bibitem[Cybenko(1989)]{Cybenko:89}
G.~Cybenko.
\newblock Approximation by superpositions of a sigmoidal function.
\newblock \emph{Mathematics of Control, Signals, and Systems}, 2\penalty0
  (4):\penalty0 303 – 314, 1989.

\bibitem[Delogu et~al.(2008)Delogu, Fanni, and Montisci]{Delogu:08}
R.~Delogu, A.~Fanni, and A.~Montisci.
\newblock Geometrical synthesis of mlp neural networks.
\newblock \emph{Neurocomputing}, 71\penalty0 (4-6):\penalty0 919 – 930, 2008.

\bibitem[Fernández-Delgado et~al.(2011)Fernández-Delgado, Ribeiro, Cernadas,
  and Ameneiro]{Fernández-Delgado:11}
M.~Fernández-Delgado, J.~Ribeiro, E.~Cernadas, and S.B. Ameneiro.
\newblock Direct parallel perceptrons (dpps): Fast analytical calculation of
  the parallel perceptrons weights with margin control for classification
  tasks.
\newblock \emph{IEEE Transactions on Neural Networks}, 22\penalty0
  (11):\penalty0 1837 – 1848, 2011.

\bibitem[Fernández-Delgado et~al.(2014)Fernández-Delgado, Cernadas, Barro,
  Ribeiro, and Neves]{Fernández-Delgado:14}
M.~Fernández-Delgado, E.~Cernadas, S.~Barro, J.~Ribeiro, and J.~Neves.
\newblock Direct kernel perceptron (dkp): Ultra-fast kernel elm-based
  classification with non-iterative closed-form weight calculation.
\newblock \emph{Neural Networks}, 50:\penalty0 60 – 71, 2014.

\bibitem[J{\'o}dar et~al.(1991)J{\'o}dar, Law, Rezazadeh, and Weston]{jodar:91}
L.~J{\'o}dar, A.G. Law, A.~Rezazadeh, and J.H. Weston.
\newblock Computations for the moore penrose and other generalized inverses.
\newblock \emph{Congressus Numerantium}, pages 57--57, 1991.

\bibitem[LeCun et~al.(2015)LeCun, Bengio, and Hinton]{lecun:15}
Y.~LeCun, Y.~Bengio, and G.~Hinton.
\newblock Deep learning.
\newblock \emph{nature}, 521\penalty0 (7553):\penalty0 436--444, 2015.

\bibitem[Marquardt(1963)]{marquardt:63}
D.~W. Marquardt.
\newblock An algorithm for least-squares estimation of nonlinear parameters.
\newblock \emph{Journal of the society for Industrial and Applied Mathematics},
  11\penalty0 (2):\penalty0 431--441, 1963.

\bibitem[Ploj(2014)]{Ploj:14}
B.~Ploj.
\newblock Optimization for multi layer perceptron: Without the gradient, 2014.

\bibitem[Ploj et~al.(2011)Ploj, Zorman, and Kokol]{Ploj:11}
B.~Ploj, M.~Zorman, and P.~Kokol.
\newblock Border pairs method - constructive mlp learning classification
  algorithm.
\newblock \emph{Lecture Notes in Computer Science (including subseries Lecture
  Notes in Artificial Intelligence and Lecture Notes in Bioinformatics)}, 6943
  LNAI:\penalty0 297 – 307, 2011.

\bibitem[Ploj et~al.(2014)Ploj, Harb, and Zorman]{Ploj:14b}
B.~Ploj, R.~Harb, and M.~Zorman.
\newblock Border pairs method-constructive mlp learning classification
  algorithm.
\newblock \emph{Neurocomputing}, 126:\penalty0 180 – 187, 2014.

\bibitem[Rumelhart et~al.(1986)Rumelhart, Hinton, and Williams]{Rumelhart:86}
D.E. Rumelhart, G.~E. Hinton, and R.~J. Williams.
\newblock Learning representations by back-propagating errors.
\newblock \emph{Nature}, 323\penalty0 (6088):\penalty0 533 – 536, 1986.

\bibitem[Schwefel(1981)]{schwefel:81}
H.-P. Schwefel.
\newblock \emph{Numerical optimization of computer models}.
\newblock John Wiley \& Sons, Inc., 1981.

\end{thebibliography}

\end{document}